\documentclass{sig-alternate}
\usepackage{amssymb,amsmath,algorithm,algorithmic}
\usepackage{url}
\newcommand{\eat}[1]{}
\usepackage{graphicx}
\newfloat{Algorithm}{thp}{}[section]
\newfloat{Algorithm}{tb}{lox}
\floatname{Algorithm}{Algorithm}
\begin{document}

\title{Transductive Classification Methods for Mixed Graphs}
\numberofauthors{2} 
\author{
\alignauthor
S Sundararajan\\
       \affaddr{Yahoo! Labs}\\
       \affaddr{Bangalore, India}\\
       \email{ssrajan@yahoo-inc.com}
\alignauthor
S Sathiya Keerthi\\
       \affaddr{Yahoo! Labs}\\
       \affaddr{Santa Clara, CA}\\
       \email{selvarak@yahoo-inc.com}
}
\conferenceinfo{MLG}{'11 San Diego, CA, USA} \CopyrightYear{2011} \crdata{978-1-4503-0834-2}
\maketitle
\begin{abstract}
In this paper we provide a principled approach to solve a transductive classification problem involving a similar graph (edges tend to connect nodes with same labels) and a dissimilar graph (edges tend to connect nodes with opposing labels). Most of the existing methods, e.g., Information Regularization (IR), Weighted vote Relational Neighbor classifier (WvRN) etc, assume that the given graph is only a similar graph. We extend the IR and WvRN methods to deal with mixed graphs. We evaluate the proposed extensions on several benchmark datasets as well as two real world datasets and demonstrate the usefulness of our ideas.
\end{abstract}

\vspace{1mm}
\noindent
{\bf Categories and Subject Descriptors:} I.5 {[Pattern Recognition]} {Design Methodology - {\it Classifier design and evaluation}}

\vspace{1mm}
\noindent
{\bf General Terms:} Algorithms, Experimentation

\vspace{1mm}

\section{Introduction}

Consider the problem of transductive classification in a relational graph consisting of labeled and unlabeled nodes. Most methods for this problem assume that connected nodes have the same labels. In many applications this assumption is violated to varying degrees depending on the underlying relational graph; that is, many edges can be formed using pairs of nodes having different class lables (this is referred to as label dissimilarity). When this happens the performance of the methods can deteriorate significantly. If such `dissimilar' edges can be identified via domain knowledge or other ways, they can be eliminated to improve the performance. Even better, it makes sense to collect the identified dissimilar edges in a {\em dissimilar graph} and use it differently but together with the {\em similar graph} (set of edges connecting nodes having same labels) to improve classification. This paper is rooted on this point. Let us refer to the combination of similar and dissimilar graphs simply as a {\em mixed graph}. Recently Goldberg et al. \cite{gold} extended the graph-based semi-supervised learning method of Sindhwani et al \cite{sindhwani} to deal with mixed graphs. In this method classification is fundamentally based on content features of nodes, with the mixed graph strongly guiding the classification. If $f_i$ and $f_j$ denote the classifier outputs associated with nodes $i$ and $j$ that form a dissimilar edge, Goldberg et al.'s method \cite{gold} includes a loss term, $(f_i+f_j)^2$ in the training objective function, thus putting pressure on $f_i$ and $f_j$ to have opposing signs. In many applications, content features are either weak or unavailable. Such problems have to be addressed in a purely graph transductive setting. In another related work, Tong and Jin~\cite{tong} proposed a graph based approach using semi-definite programming (SDP) to explore both similar and dissimilar graphs. The problem solved in their work is a non-convex programming problem whose solution can lead to local optima. In contrast the proposed methods in this paper are simpler and more efficient. Further our extension of the information regularization method for mixed graph (IR-MG) leads to a convex programming problem and the proposed algorithm converges to the global solution.  

The main aim of this paper is to extend and explore existing methods for a transductive setting to deal with mixed graphs (even when non-content based relational graphs are available). We only take up binary classification in this paper. There are many worthy methods in this group of methods; examples are: Information Regularization (IR) \cite{cord}, Weighted vote Relational Neighbor classifier (WvRN) \cite{sofusjmlr}, Local and Global Consistency (LGC) \cite{zhou} and Gaussian Function Harmonic Field (GFHF) \cite{zhu}. To keep the paper short we take only the first two methods for extension. Both these methods are based on probabilistic ideas; thus, instead of the squared loss used by Goldberg et al. \cite{gold}, we devise a divergence-based convex loss function to deal with dissimilar edges. Empirical results show that the extensions are very effective, although the ideas are simple and straight-forward.

Depending on the way they are formed, the similarity and dissimilarity graphs in a given problem may differ in pureness. So it is useful to have a hyperparameter ($\gamma$) that mixes the effects of these two graphs (e.g., relative weighting between the losses corresponding to the two graphs). We make use of such a parameter; our experiments on the various datasets point to the importance of this parameter. Though Goldberg et al. \cite{gold} do not use such a parameter, it appears to be useful for that method too. The quality of the graphs relating to classification solution can also be approximately measured using a quantity called {\em node assortativity coefficient} (NAC) \cite{sofusaaai}. NAC is easy to compute and gives a good indication of the usefulness of the graphs for classification. It can also be used to quickly select a decent value for the $\gamma$ parameter.

To demonstrate the effectiveness of our extended methods we do detailed experiments, like Goldberg et al. \cite{gold}, on standard academic benchmark datasets in which mixed graphs are constructed systematically but artificially. We also show usefulness of our methods on real world datasets involving web pages of shopping domains. In these problems mixed graphs arise naturally. For example, two web pages that either have strong structural similarity or have co-citation links from a common third page may have the same labels, and, web pages that have extremely poor structural correlation may have opposing labels.

The paper is organized as follows. In section 2 we give the extensions of IR and WvRN for mixed graphs. In section 3 we define NAC and discuss its usefulness; hyperparameter tuning is also discussed there. Experimental results are given in section 4 and we conclude with section 5.

The following notations will be used in this paper. Let ${\bf G}~=~({\bf V},{\bf E},{\bf W})$ be an undirected graph with ${\bf V}~=~ \\ \{v_1,\cdots,v_n\}$ representing the set of nodes, ${\bf E}$ and ${\bf W}$ representing the set of edges and associated weights respectively. Assume that $w_{i,j}\ge 0$, $\forall i,j$ where $w_{i,j}$ represents the edge weight between the nodes $v_i$ and $v_j$. In a graph ${\bf G}$ typically we have both similar and dissimilar edges. Similar edges connect nodes belonging to same class and dissimilar edges connect nodes belonging to different classes. Since an edge can be either similar or dissimilar we can separate the graph ${\bf G}$ into {\it similar} and {\it dissimilar} graphs (denoted as ${\bf G}_S$ and ${\bf G}_{\bar S}$) respectively. Then the nodes, edges and weights corresponding to these graphs are appropriately defined as: ${\bf G}_{S}=~({\bf V}_S,{\bf E}_S,{\bf W}_S)$ and ${\bf G}_{\bar S}=~({\bf V}_{\bar S},{\bf E}_{\bar S},{\bf W}_{\bar S})$. Let ${\bf p}_i$ and ${\bf q}_i$ denote two probability distributions over the set of possible labels, associated with the node $v_i$. Usually ${\bf p}_i$ represents any known or a prior distribution for node $v_i$ and ${\bf q}_i$ represents probability distribution estimate obtained from any given method. In this paper we are interested only in binary classification problem and so ${\bf p}_i$ and ${\bf q}_i$ are 2-dimensional vectors. Also, let ${\bf P}=[{\bf p}_1,\ldots,{\bf p}_n]$ and ${\bf Q}=[{\bf q}_1,\ldots,{\bf q}_n]$. Let $L$ and $UL$ denote the set of labeled and unlabeled nodes respectively. 

\section{Proposed Methods}
In this section we show how two existing methods, namely, information regularization (IR) and Weighted vote Relational Neighbor classification (WvRN) can be extended to handle the mixed graph scenario. 
\subsection{Information Regularization in a mixed \\ graph setting}
In the conventional setting only similar edges are {\it assumed}. That is, we have ${\bf G}={\bf G}_S$ and the edge weights $w_{i,j}, \forall (i,j) \in {\bf E}$ in some sense indicate our belief or confidence in the assumption that the connected nodes belong to same class. Within that assumption, we consider solving the transductive classification problem by optimizing the objective function:
\begin{equation}
	F({\bf Q};{\bf P},{\bf W})=\sum_{i \in L} {\cal D}({\bf p}_i || {\bf q}_i)+ \\ \lambda_S \sum_{(i,j) \in E}w_{i,j} {\cal D}({\bf q}_i || {\bf q}_j)
\label{eq1}
\end{equation}
where ${\cal D}(\cdot)$ denotes any divergence measure that measures the dissimilarity between two distributions. Several divergence measures have been used in the literature. They include Kullback-Leibler (KL)
divergence, Jensen-Shannon (JS) divergence, Jensen-Renyi (JR) divergence etc. Here, we consider Jensen-Shannon divergence which is a symmetric and smoothed version of the KL divergence. When ${\cal D}(\cdot)$ is taken as the JS-divergence the regularization term is nothing but the {\it information regularization} proposed by Corduneanu and Jaakkola \cite{cord} in a graph setting. The first term in (\ref{eq1}) is a data fitting term and measures how well the estimate ${\bf q}_i$ matches the input distribution ${\bf p}_i$, $\forall i \in L$. The second term is a regularization term and it regularizes the solution ${\bf Q}^*$ with respect to the underlying relational graph. The regularization constant $\lambda_S$ trades off between the data fitting and regularization terms. When two nodes are strongly connected their distributions are expected to be similar and the regularization term enforces this behavior. Clearly, if the individual terms are convex then the solution is unique.  

(\ref{eq1}) assumes that all the edges are similarity edges (i.e., ${\bf E}={\bf E}_S$). Therefore depending on the extent to which this assumption is violated the performance suffers. To address this problem we propose the following modified objective function: 
\begin{equation}
\begin{split}
F({\bf Q};{\bf P},{\bf W}) & = \sum_{i \in L} {\cal D}({\bf p}_i || {\bf q}_i)+\lambda_S \sum_{(i,j) \in {\bf E}_S}w_{i,j} {\cal D}({\bf q}_i || {\bf q}_j) \\
& \quad \lambda_{\bar S}\sum_{(i,j) \in {{\bf E}_{\bar S}}}w_{i,j} {\cal D}({\bf q}_i || {\bf H}_{1,2}{\bf q}_j)
\end{split}
\label{eq2}
\end{equation}
where ${\bf H}_{1,2}=\begin{pmatrix} 0 & 1 \\ 1 & 0 \end{pmatrix}$ is a transformation matrix and $\lambda_{\bar S}$ is another regularization constant. Let ${\tilde{\bf q}}_j={\bf H}_{1,2}{\bf q}_j$. Clearly ${\tilde {\bf q}}_j={\bf 1}-{\bf q}_j$ is still a distribution and the transformation facilitates divergence measurement of the distributions ${\bf q}_i$ with the distributions ${\bf 1}-{\bf q}_j$ for the edges in ${\bar S}$. Here, ${\bf 1}$ represents a vector of all ones. Therefore the dissimilar edges will also help in reinforcing the class distributions in a positive way.   

Corduneanu and Jaakkola \cite{cord} proved that the solution to (\ref{eq1}) with information regularization is unique. Using the constraint ${\bf q}_i~={\bf p}_i$, $\forall i \in L$ they suggested a distributed propagation algorithm that finds the solution in an iterative fashion. In a similar way one can show that (\ref{eq2}) is also convex and that the solution can be found in an iterative fashion. The proof is based on a standard log sum inequality and properties of KL-divergence measure \cite{cover}. Therefore (\ref{eq2}) is a natural extension of the information regularization approach in the mixed graph setting; we will refer to this method as information regularization for mixed graphs (IR-MG). The algorithm is given in \ref{alg:IR-MG}. 

We note that when we set ${\bf q}_i~={\bf p}_i$, $\forall i \in L$ and optimize ${\bf q}_i$ {\it only} for $i \in UL$ then the solution ${\bf Q}^*_{UL}$ is dependent only on the second term in (\ref{eq1}), and, second and third terms in (\ref{eq2}). Such a setting is useful when the labels are clean and the graph is not extremely dense in some regions~\cite{subra}. Both these requirements can be often met in many practical applications. When they cannot be met, methods proposed in~\cite{subra} are useful to solve (\ref{eq1}). Such methods can be appropriately extended to find the solution for our problem of mixed graphs.   

In a normalized graph setting, one way to normalize is to do node level normalization using its degree separately in each graph. That is, set ${\bf W}_S={\bf D}^{-1}_{S}{\bf W}_{S}$ and ${\bf W}_{\bar S}={\bf D}^{-1}_{\bar S}{\bf W}_{\bar S}$ where $[{\bf D}_S]_{ii}=\sum_{j|(i,j) \in {\bf E}_S)} w_{i,j}$ and $[{\bf D}_{\bar S}]_{ii}= \\ \sum_{j|(i,j) \in {\bf E}_{\bar S}} w_{i,j}$. Then, we can set $\lambda_S=\lambda \gamma$ and $\lambda_{\bar S}=\lambda (1-\gamma)$ where $\lambda>0$ and $0\le \gamma \le 1$. In such a case, $\lambda$ is the overall regularization constant and $\gamma$ weighs the similar and dissimilar contributions. When we set ${\bf q}_i~={\bf p}_i$, $\forall i \in L$, we have only one parameter $\gamma$. In practice since the graphs are {\it impure} (i.e., it may not be possible to construct {\it pure} similar and dissimilar graphs) to varying degrees, the $\gamma$ parameter plays an important role in achieving improved performance.    

\begin{Algorithm}
	\caption{IR-MG Algorithm}
	\label{alg:IR-MG}
\begin{algorithmic}
	\STATE $t \leftarrow 0$ and $\epsilon=0.001$
	\STATE For all nodes $i \in {\it UL}$, initialize ${\bf q}^{(t)}_i$ to the class prior (obtained from known labeled nodes) and {\it fix} ${\bf q}_i={\bf p}_i$ $\forall i \in {\it L}$. 
	\REPEAT 
              \FOR{each edge $(i,j)\in {\bf E}_S$}  
              \STATE ${\bf u}_{i,j} \leftarrow 0.5({\bf q}^{(t)}_i+{\bf q}^{(t)}_j)$ 
              \ENDFOR
              \FOR{each edge $(i,j)\in {\bf E}_{\bar S}$} 
              \STATE ${\bf z}_{i,j} \leftarrow 0.5({\bf q}^{(t)}_i+1-{\bf q}^{(t)}_j)$ 
              \ENDFOR
	      \FOR{each element $i \in {\it UL}$} 
              \STATE ${\bf q}^{(t+1)}_{i} \leftarrow \frac{1}{\psi} \exp(\gamma \sum_{j|(i,j)\in {\bf E}_S} w_{i,j} \log({\bf u}_{i,j})+(1-\gamma)\sum_{j|(i,j)\in {\bf E}_{\bar S}} w_{i,j} \log({\bf z}_{i,j}))$
        \ENDFOR
        \STATE $t \leftarrow t+1$
 \UNTIL {$\max_{i \in {\it UL},k=1,2} |q^{(t-1)}_{i,k}-q^{(t)}_{i,k}| < \epsilon$}
\end{algorithmic}
\end{Algorithm}
\subsection{WvRN Classification in a mixed graph setting}
The original probabilistic Weighted vote Relational Neighbor classifier (with relaxation labeling) method \cite{sofusjmlr} was formulated to solve the collective classification problem (for {\it only} similar graphs) where class distributions of a subset of nodes are known and fixed. Then the class distributions of the remaining (unlabeled) nodes are obtained by an iterative algorithm. It has two components, namely, weighted vote relational neighbor classifier component and relaxation labeling (RL) component. The relaxation labeling component performs collective inferencing and keeps track of the current probability estimates ${\bf q}^{(t)}_i$ for all unlabeled nodes at each time instant $t$. These {\it frozen} estimates ${\bf q}^{(t)}_i$ are used by the relational classifier. The relational classifier computes the probability distribution for each unlabeled node as the weighted sum probability distributions ${\bf q}^{(t)}_j$ of its neighbors with weight $w_{ij}$; that is, 
\begin{equation}
              q^{(t+1)}_{i,k}=\frac{1}{\psi}\sum_{j} w_{i,j} q^{(t)}_{j,k}
\label{WvRN1}
\end{equation}
where $k=1,2$ and $\psi$ is a normalizing constant. Since relaxation labeling may not converge, sometimes simulated annealing is performed to ensure convergence~\cite{sofusjmlr}.

In a mixed graph setting, we can modify (\ref{WvRN1}) as:
\begin{equation}
              q^{(t+1)}_{i,k}=\frac{1}{\psi}(\gamma\sum_{j|(i,j)\in {\bf E}_S} w_{i,j} q^{(t)}_{j,k}+(1-\gamma)\sum_{j|(i,j)\in {\bf E}_{\bar S}} w_{i,j} (1-q^{(t)}_{j,k}))
\label{WvRN2}
\end{equation}
where $k=1,2$. As in the case of IR-MG method, the parameter $\gamma$ weighs the similar and dissimilar graphs. With the modification given in (\ref{WvRN1}), we refer to this method as WvRN-MG. The algorithm is given in \ref{alg:WvRN-MG}.       

\begin{Algorithm}
	\caption{WvRN-MG Algorithm}
	\label{alg:WvRN-MG}
\begin{algorithmic}
	\STATE $t \leftarrow 0$, $\beta^{(t)} \leftarrow 1$, $\nu \leftarrow 0.95$ and $\epsilon=0.001$
	\STATE For all nodes $i \in {\it UL}$, initialize ${\bf q}^{(t)}_i$ to the class prior (obtained from known labeled nodes) and {\it fix} ${\bf q}_i={\bf p}_i$ $\forall i \in {\it L}$. 
	\REPEAT
	      \FOR{each element $i \in {\it UL}$ and $k=\{1,2\}$}
              \STATE ${\tilde q}_{i,k} \leftarrow \frac{1}{\psi}(\gamma\sum_{j|(i,j)\in {\bf E}_S} w_{i,j} q^{(t)}_{j,k}+(1-\gamma)\sum_{j|(i,j)\in {\bf E}_{\bar S}} w_{i,j} (1-q^{(t)}_{j,k}))$
			\STATE $q^{(t+1)}_{i,k} \leftarrow \beta^{(t)}{\tilde q}_{i,k}+(1-\beta^{(t)})q^{(t)}_{i,k}$
         \ENDFOR
              \STATE $t \leftarrow t+1$ and $\beta^{(t+1)} \leftarrow \beta^{(t)}*\nu$
    \UNTIL {$\max_{i \in {\it UL},k=1,2} |q^{(t-1)}_{i,k}-q^{(t)}_{i,k}| < \epsilon $}
\end{algorithmic}
\end{Algorithm}
\section{Graph Characteristics and Setting $\gamma$}
Characteristics of graphs play a major role in achieving good classification performance. One of the key characteristics of a relational graph is the correlation of the class variable of related entities. A graph is said to have {\it homophily} characteristics when the related entities in the graph have the same label; this was studied by early social network researchers. All the methods that make use of this assumption are essentially homophily based methods \cite{sofusjmlr}. There is also a link-centric notion of homophily known as {\it assortativity} studied in \cite{newman}. The {\it assortativity coefficient} \cite{newman} measures the homophily characteristics based on the correlation between the classes linked by edges in the graph. Macskassy and Provost \cite{sofusjmlr} developed a variant of this coefficient. It is based on the graph's node assortativity matrix ${\bf C}$ where $C_{ij}$ represents, for all nodes of class $y_i$,  the average weighted fraction of their weighted edges that connect them to nodes of class $y_j$ such that $\sum_{i,j}C_{ij}=1$. Then the node assortativity coefficient (NAC) $N$ is defined as: $N={{\sum_{i}C_{ii}-\sum_i a_i.b_i} \over {1-\sum_i a_i.b_i}}$ where $a_i$ and $b_i$ denote the sum of the $i$-th row and $i$-th column respectively. This coefficient takes values in [-1,1] with the extremes indiciating strong connectivity between dissimilar and similar classes respectively. Macskassy and Provost \cite{sofusjmlr} used this coefficient to study its usefulness in edge selection \cite{sofusjmlr}. Macskassy \cite{sofusaaai} used this coefficient to weigh different edge types when there are multiple graphs. Specifically, each edge was scaled by its graph's $N$ value; if it is negative the scaling factor for that edge type (graph) was set to zero. Since the original WvRN is a homophily based method, Macskassy and Provost \cite{sofusjmlr} set the weight to zero for graphs having negative $N$ values. We illustrate below how this coefficient can be used to set $\gamma$ in the mixed graph scenario.

In our proposed methods, the mixture parameter $\gamma$ plays an important role since it decides the degree to which each graph controls the performance. In practice this parameter can be set in two ways. One way is to set $\gamma$ using the NAC values of similar and dissimilar graphs. Let $N_S$ and $N_{\bar S}$ denote estimates of the NAC values of the similar and dissimilar graphs respectively. Note that, if the dissimilar graph is pure (for example, as in section 4.1 below) then $N_{\bar S}=-1$. Therefore, we can set $\gamma={N_S\over{N_S-N_{\bar S}}}$. If $N_S<0$ and/or $N_{\bar S}>0$ it is not a good idea to use the above estimate of $\gamma$.
For best performance it is a good idea to set $\gamma$ using cross-validation. However, unlike NAC based $\gamma$ estimation, the CV technique is expensive since we need to run the training algorithm several times. Finally, note that, since both methods are based on labeled nodes, a good estimate of $\gamma$ can be obtained only when the number of labeled nodes is not too small. In section 4.3 we illustrate the usefulness of these techniques on several benchmark datasets.
\begin{table*}
\begin{center}
\caption{Properties of datasets: $n$ and $e$ denote the number of nodes and edges in ${\bf G}$ respectively; $L$, $n_f$, $b$ and $N_s$ denote the number of labeled nodes, the number of (content) features, percentage of positive examples and node assortativity coefficient values respectively.}
\label{tab:properties}
\vskip 0.01in
\begin{tabular}{|l|r|r|r|r|r|r|}\hline
Dataset & \multicolumn{1}{|c}{$n$} & \multicolumn{1}{|c}{$e$} & \multicolumn{1}{|c}{$n_f$}& \multicolumn{1}{|c}{$b$}&  \multicolumn{1}{|c}{$N_s$} & \multicolumn{1}{|c|}{$L-Range$}\\ \hline
G50C  &  550  &   40940   &   550 & 50  &   0.47   & 10-50\\ \hline
WINDOWS-MAC  &   1946  &   124806   &   7511 & 50.62 &  0.49  & 50-400\\ \hline
WebKB-PAGELINK   & 1051  &  269044  & 4840 &  21.88  & 0.57 & 10-50  \\ \hline
WebKB-LINK   & 1051  &  72446   &  1840 & 21.88  & 0.55   & 10-50 \\ \hline
IMDBALL   & 1441  &  48371   &  - & 57.32  & 0.36   & 50-400 \\\hline
CORAALL1   & 4240  &  71802   &  - & 6.25  & 0.69   & 50-800 \\ \hline
CORAALL2   & 4240  &  71802   &  - & 8.28  & 0.59   & 50-800 \\ \hline
CORAALL3   & 4240  &  71802   &  - & 12.33  & 0.60   & 50-800 \\ \hline
CORAALL4   & 4240  &  71802   &  - & 32.17  & 0.67   & 50-800 \\ \hline
UG-Product (${\bf G}_{S}$) & 1166  & 54462 & - & 39.71 & 0.99 & 40-160 \\
UG-Product (${\bf G}_{\bar S}$) & 1166  & 47327 & - & 39.71 & 0.76 & 40-160 \\ \hline
UG-Listing (${\bf G}_{S}$) & 1166  & 54462 & - & 54.55 & 0.96 & 40-160 \\
UG-Listing (${\bf G}_{\bar S}$) & 1166  & 47327 & - & 54.55 & 0.52 & 40-160 \\ \hline
CU-Product (${\bf G}_{S}$) & 1433  & 26201 & - & 46.55 & 0.44 & 40-160 \\
CU-Product (${\bf G}_{\bar S}$) & 1433  & 71650 & - & 46.55 & 0.23 & 40-160 \\ \hline
CU-Listing (${\bf G}_{S}$) & 1433  & 26201 & - & 35.03 & 0.96 & 40-160 \\
CU-Listing (${\bf G}_{\bar S}$) & 1433  & 71650 & - & 35.03 & 0.46 & 40-160 \\ \hline
\end{tabular}
\end{center}
\end{table*}

\section{Experiments}
In this section we present results obtained from various experiments conducted on several academic benchmark datasets as well as real world datasets formed from web pages of shopping web sites. First we study the performances of the proposed methods, namely, IR-MG and WvRN-MG on mixed graphs constructed from already available relational graphs of benchmark datasets; these results demonstrate gains that accrue as a result of moving from a noisy similar graph towards a quite pure similar-dissimilar graph combination. Next, we evaluate the performances on similar and dissimilar graphs that arise naturally from web pages of shopping sites. Finally, we compare the relative performances of our methods as well as evaluate them against the method of Goldberg et al.\cite{gold}.

\subsection{Experiments on partitions of given graph into dissimilar and similar graphs}
 Usually a given relational graph (${\bf G}$) with partially labeled nodes is impure and consists of both similar and dissimilar edges. For our experiments we {\it extract} similar and dissimilar graphs (denoted as ${\bf G}_S$ and ${\bf G}_{\bar S}$) from ${\bf G}$ using the following model. Similar to the work of Goldberg et al. \cite{gold} we use an oracle which takes a pair of nodes and tells whether the edge formed by them is similar or dissimilar. We construct ${\bf G}_{\bar S}$ by randomly picking {\it a percentage of dissimilar edges} ($P$) connecting {\it only unlabeled nodes} in ${\bf G}$ by querying the oracle. Note that the learner only knows that the edges are dissimilar; it does not know the actual labels of the nodes. Thus, the dissimilar graph is a {\it pure} graph consisting of only unlabeled nodes. Then the {\it similar} graph ${\bf G}_S$ is obtained as ${\bf G}$ - ${\bf G}_{\bar S}$. Note that, unlike ${\bf G}_{\bar S}$, ${\bf G}_S$ may not be pure. This is because we vary the percentage of edges picked from ${\bf G}$ to construct ${\bf G}_{\bar S}$; also, even if we pick all the dissimilar edges connecting unlabeled nodes, there can still be some dissimilar edges connecting labeled and unlabeled nodes left in ${\bf G}_{S}$. This model is different from the model used by Goldberg et al. \cite{gold}. In that work, the original graph ${\bf G}$ is taken as ${\bf G}_{S}$ and, ${\bf G}_{\bar S}$ is constructed by taking random pairs of nodes having opposing labels using the oracle. Our model is appropriate when we are given a graph and there is some way of filtering out dissimilar edges from it. On the other hand, the model used by Goldberg et al. \cite{gold} is appropriate when we are given a similar graph and, additionally one can construct a dissimilar graph using domain knowledge. In both models the dissimilar graph is pure; one can also think of experimenting with alternate models which introduce some noise in the dissimilar graph.

A summary description of various benchmark datasets used in the experiments is given in Table \ref{tab:properties}. All the datasets indicated correspond to binary classification problems. The datasets G50C, WINDOWSMAC, WebKB-PAGELINK and WebKB-LINK used in \cite{sindhwani} are taken from \url {http://people.cs.uchicago.edu/~vikass/research.html}. G50C is an artificial dataset generated from two unit covariance normal distributions with equal probabilities; the means are adjusted so that the true Bayes error is $5\%$~\cite{sindhwani}. WINDOWSMAC dataset is a subset of 20-newsgroup dataset with the documents belonging to two categories {\it windows} and {\it mac}. The WebKB dataset arises from hypertext-based categorization of web documents with two classes {\it course} and {\it non-course}. The WebKB-LINK dataset uses features derived from the anchortext associated with links on other webpages that point to a given web page. The WebKB-PAGELINK dataset uses both PAGE and LINK features where PAGE features are derived from the content of a page. In each of these four datasets mentioned above, 
following~\cite{sindhwani,gold}, we construct the relational graph with $k$-nearest neighbors using Gaussian weights. Specifically, the weight between kNN points $x_i$ and $x_j$ is $e^{-\frac{||x_i-x_j||^2}{2\sigma^2}}$, while other weights are zero; $k$ is set to $50$, $10$ and $200$ for G50C, WINDOWSMAC and WebKB datasets respectively. 
We also consider the datasets, CORAALL and IMDBALL that do not have any input feature representation. They have the relational graph matrix ${\bf W}$ constructed purely from underlying relations. The CORAALL dataset is derived from the CORA dataset which comprises of computer science research papers; the relational graph is constructed using {\it both} co-citation and common author relationships between papers. This dataset has seven classes with each class representing topics like {\it Neural Networks}, {\it Genetic Algorithms} etc. We converted this seven class problem into 7 {one versus all} binary classification problems and the corresponding datasets are referred as CORAALL1, CORAALL2 and so on, with the number indicating the positive class. The IMDBALL dataset is based on networked data from the Internet Movie Database (IMDb) (\url{http://www.imdb.com}); here classification is about predicting movie success determined by box-office receipts (high-revenue versus low-revenue) and the relational graph is constructed between movies by linking them when they share a production company. The weight of an edge in the resulting graph is the number of production companies two movies have in common~\cite{sofusjmlr}. The CORAALL and IMDBALL datasets are available with the toolkit described in \cite{sofusjmlr}.

Next we give more details on the experiments. We provide plots only for a few datasets and comment on other datasets when needed. For each dataset, we varied the number of labeled nodes ($L$), the mixture parameter $\gamma$ and the percentage of dissimilar edges ($P$) in ${\bf G}$ used for forming the dissimilar graph. In all our experiments we considered 25 realizations where each realization corresponds to one random stratified labeling of nodes.

\begin{figure*}
\centering
        \includegraphics[width=7cm,height=3.5cm]{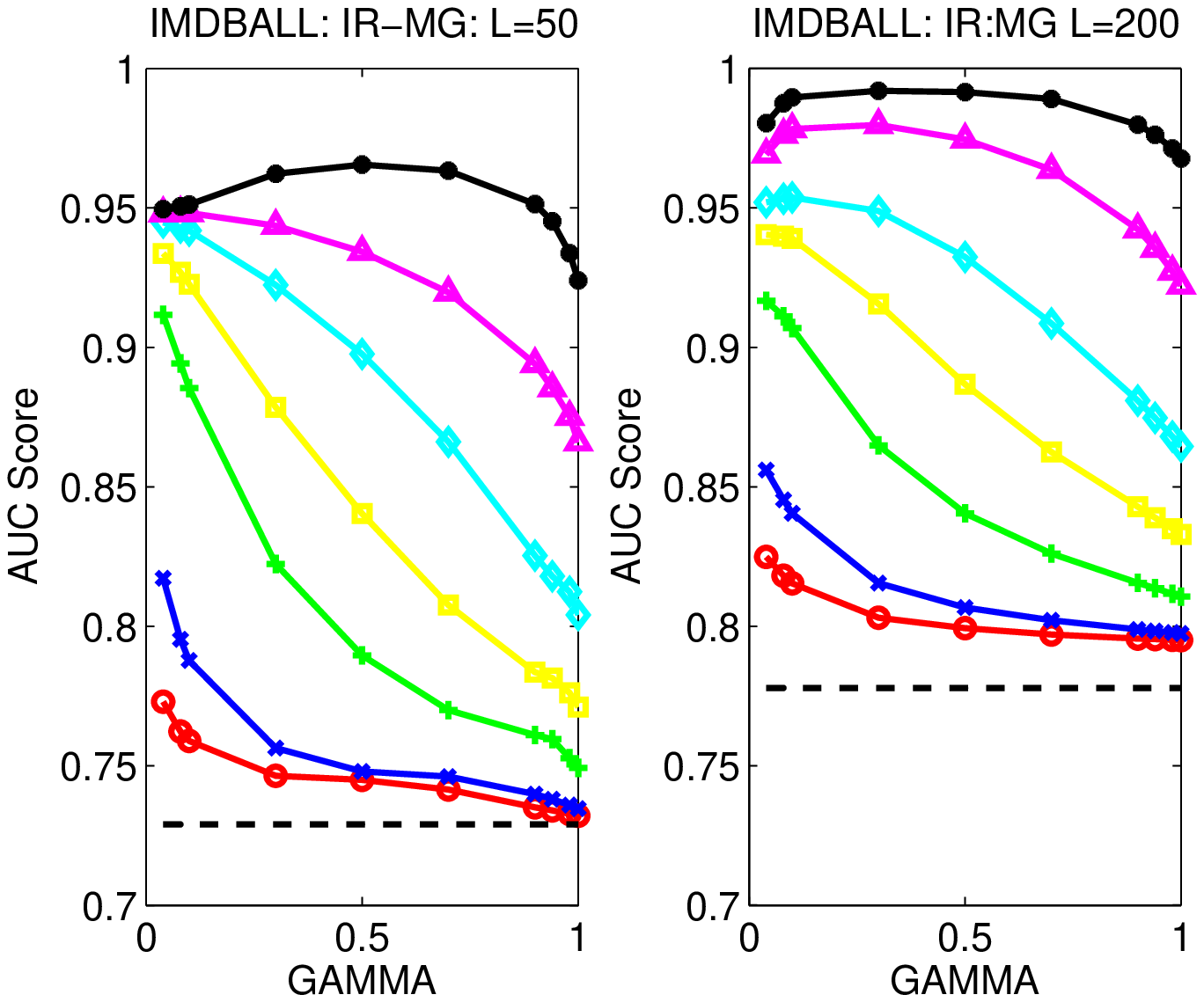}
        \includegraphics[width=7cm,height=3.5cm]{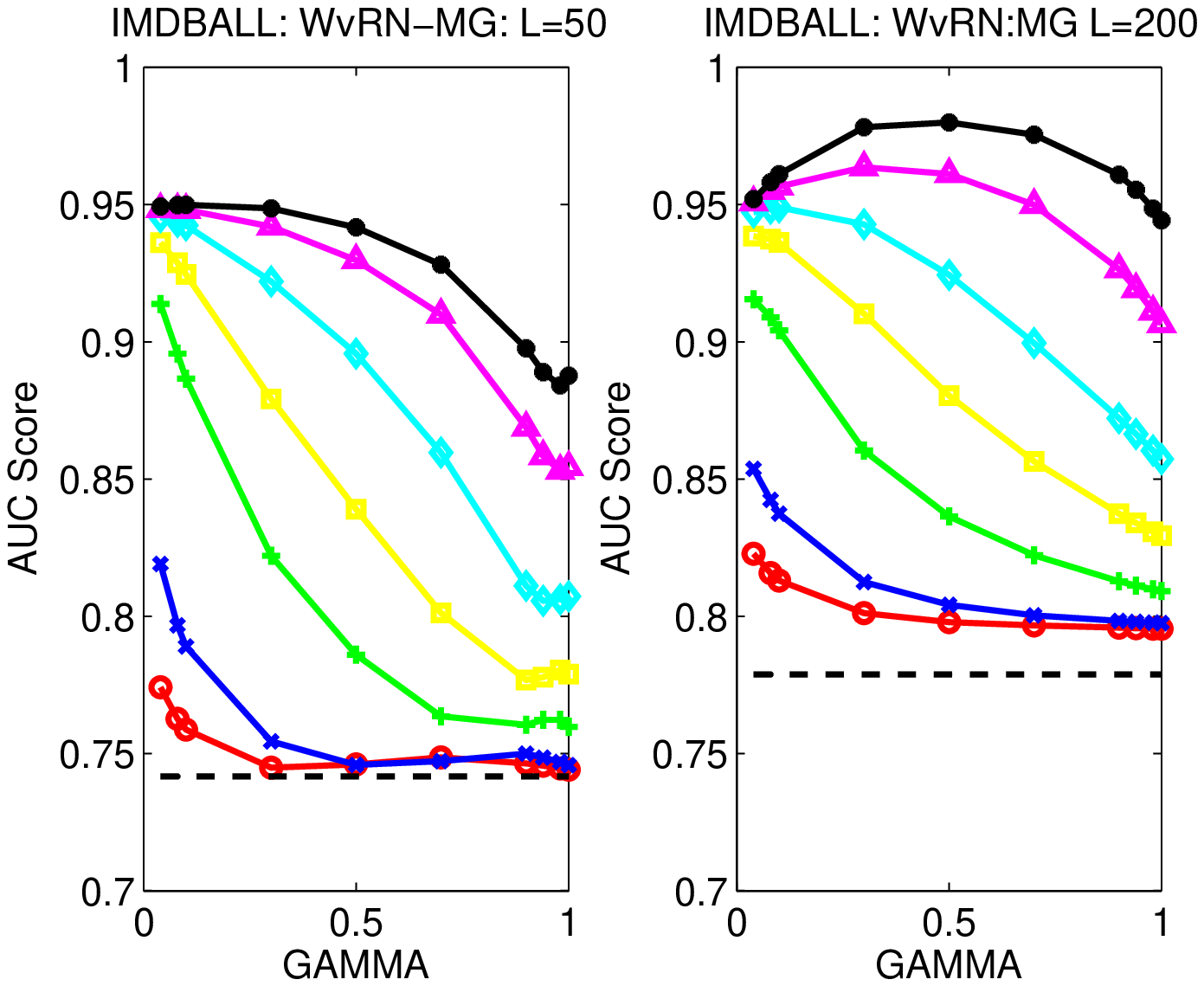}\\
        \includegraphics[width=7cm,height=3.5cm]{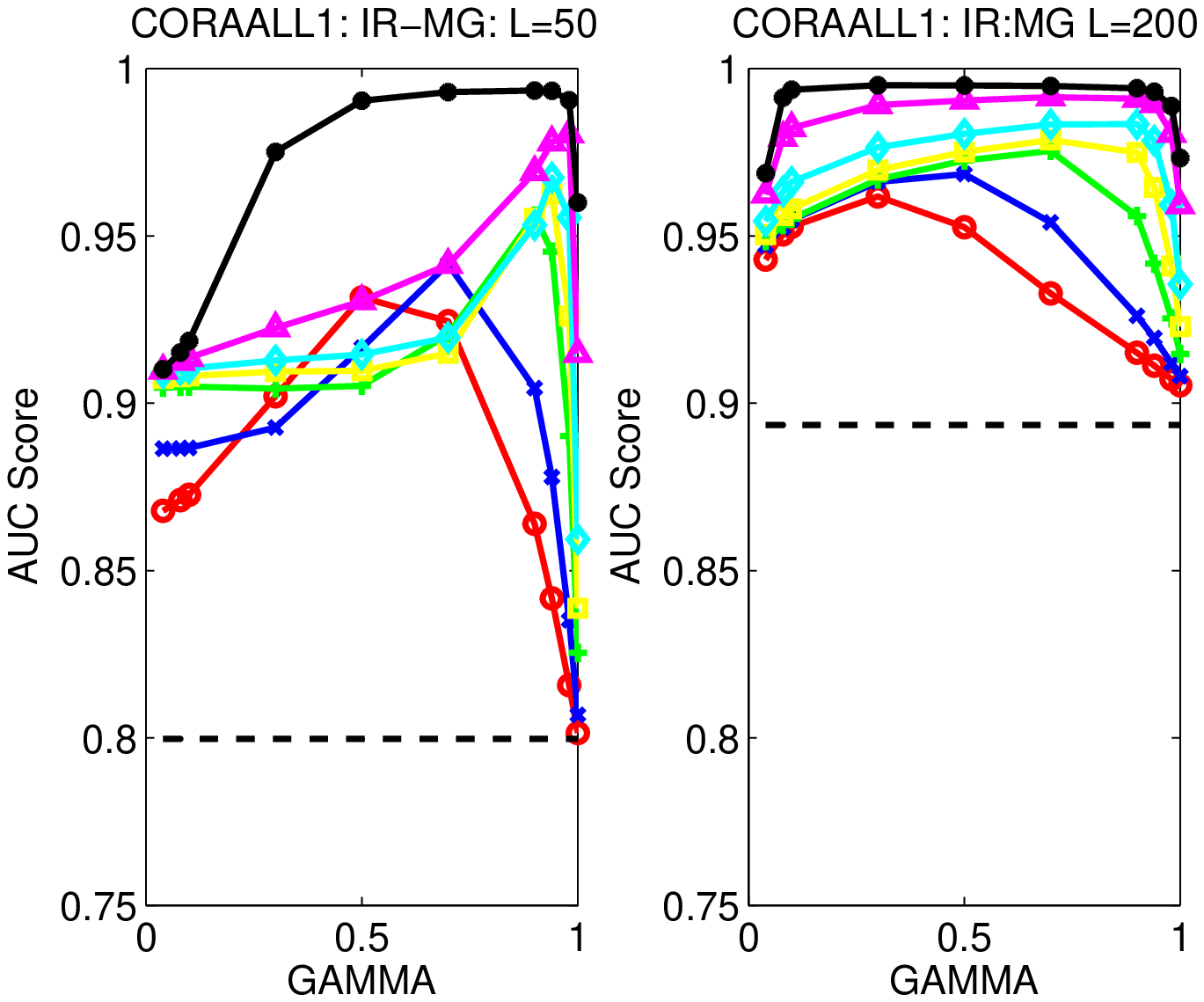}
        \includegraphics[width=7cm,height=3.5cm]{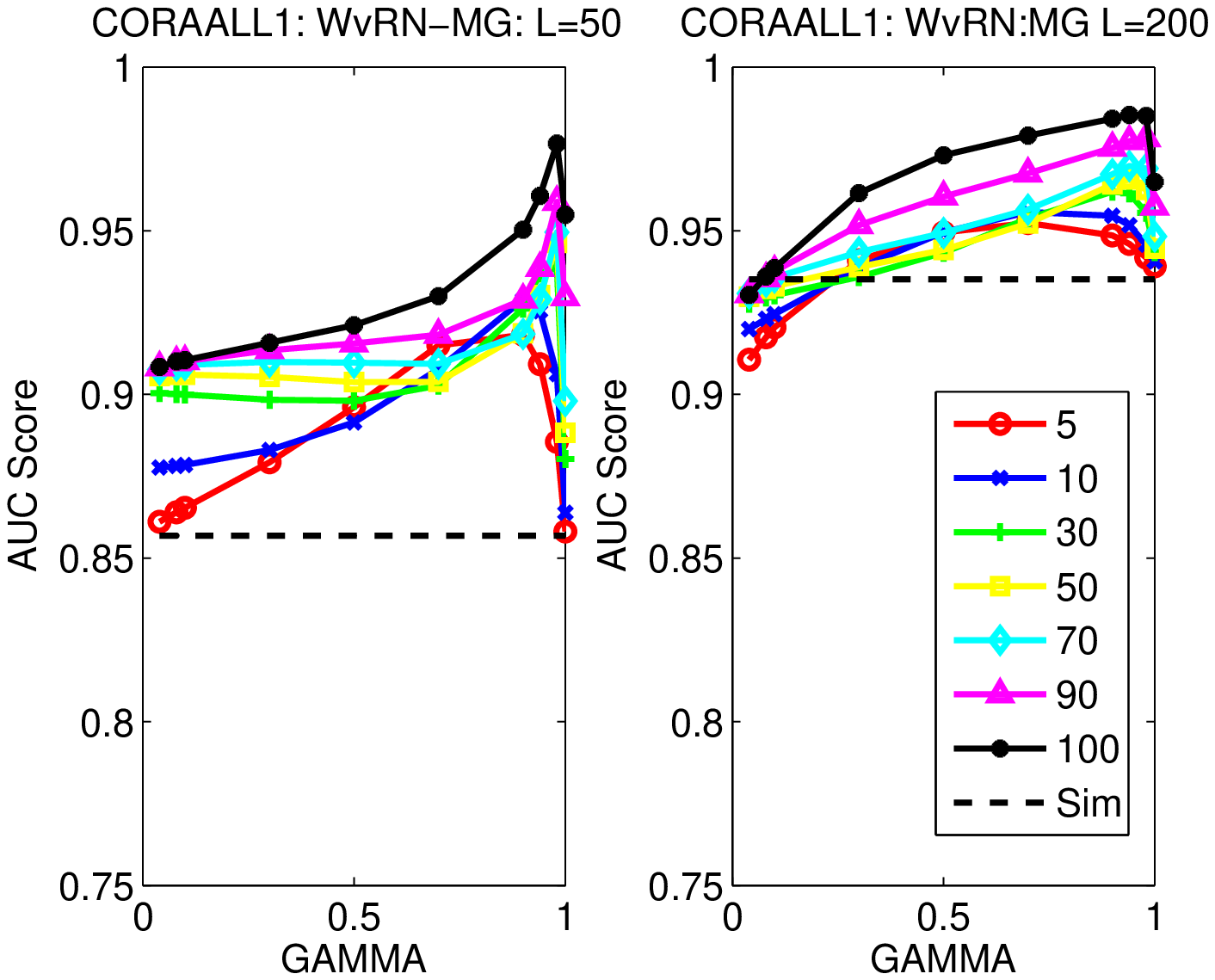}\\
\caption{{AUC score performance the of IR-MG and WvRN-MG methods on IMDBALL and CORAALL1 datasets under two different label size conditions. The numbers in the legend (applicable for all plots) indicate the percentage of dissimilar edges (with respect to the total number of dissimilar edges connecting unlabeled nodes) in ${\bf G}_{\bar S}$. The dotted black line indicate the performance with the original graph ${\bf G}$.
}} \label{fig1}
\end{figure*}

We present various observations from the experimental study conducted on all the academic benchmark datasets given in Table \ref{tab:properties}. Compared to using the original graph ${\bf G}_S$ significant performance improvements were observed with the use of the mixed graph, in a vast majority of cases of varying $P$, $\gamma$ and $L$ on all the datasets. Performance results on two representative datasets, viz. IMDBALL and CORAALL1 are given in figure \ref{fig1}. It is clearly seen that the best performance is achieved for some intermediate values of $\gamma$; see for instance the results of CORAALL1, IR-MG, L=80 and 200. This demonstrates that although the similar graph is noisy, it is still useful in the mixed graph setting to get improved performance. In the case of IMDBALL dataset, the best performance is achieved at low value of $\gamma$ and smaller $P$ values; this is because the similar graph is more noisy (with the original graph having a node assortativity coefficient of only 0.36). However, for large $P$ values, the similar graph becomes purer (but still noisy) and the best performance is achieved again for some intermediate values of $\gamma$. 

We also conducted paired-t statistical significance tests to compare IR-MG and WvRN-MG methods on each dataset. On the original graph, the WvRN-MG method was slighly better on WebKB-PAGELINK, CORAALL1, CORAALL2, CORAALL3 and CORAALL4 datasets and the significance reduces as the number of labeled nodes is increased. Next we consider the mixed graph case. In the case of CORAALL1 dataset, we observed that the IR-MG method started performing better in an intermediate range of values of $\gamma$ as the graph becomes purer. 
At higher $\gamma$ values (corresponding to the original graph when $P=0$ and subsequently purer similar graph as $P$ increases), there was no statistical significance found. Similar observations were found in the case of IMDBALL dataset. 
Overall we found that the IR-MG method performs better on purer graphs. 


In practice we need automatic ways of using domain knowledge or otherwise to identify similar and dissimilar edges. This is an important research topic; but it is beyond the scope of this paper. In several applications similar and dissimilar graphs occur naturally, and both the  graphs are typically noisy. We demonstrate the usefulness of the proposed methods on one such application next.
\begin{figure*}
\centering
        \includegraphics[width=3.5cm,height=3.5cm]{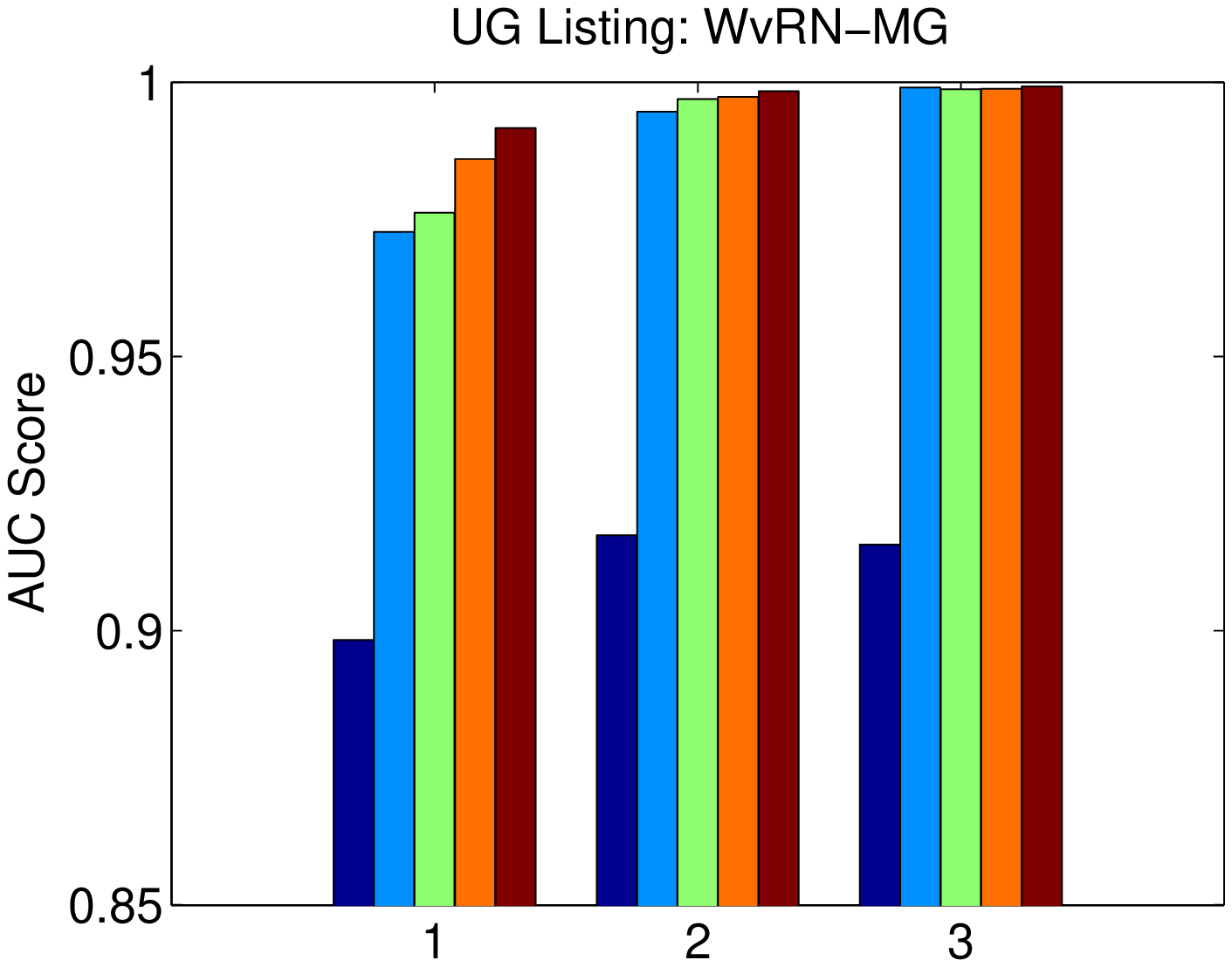} 
        \includegraphics[width=3.5cm,height=3.5cm]{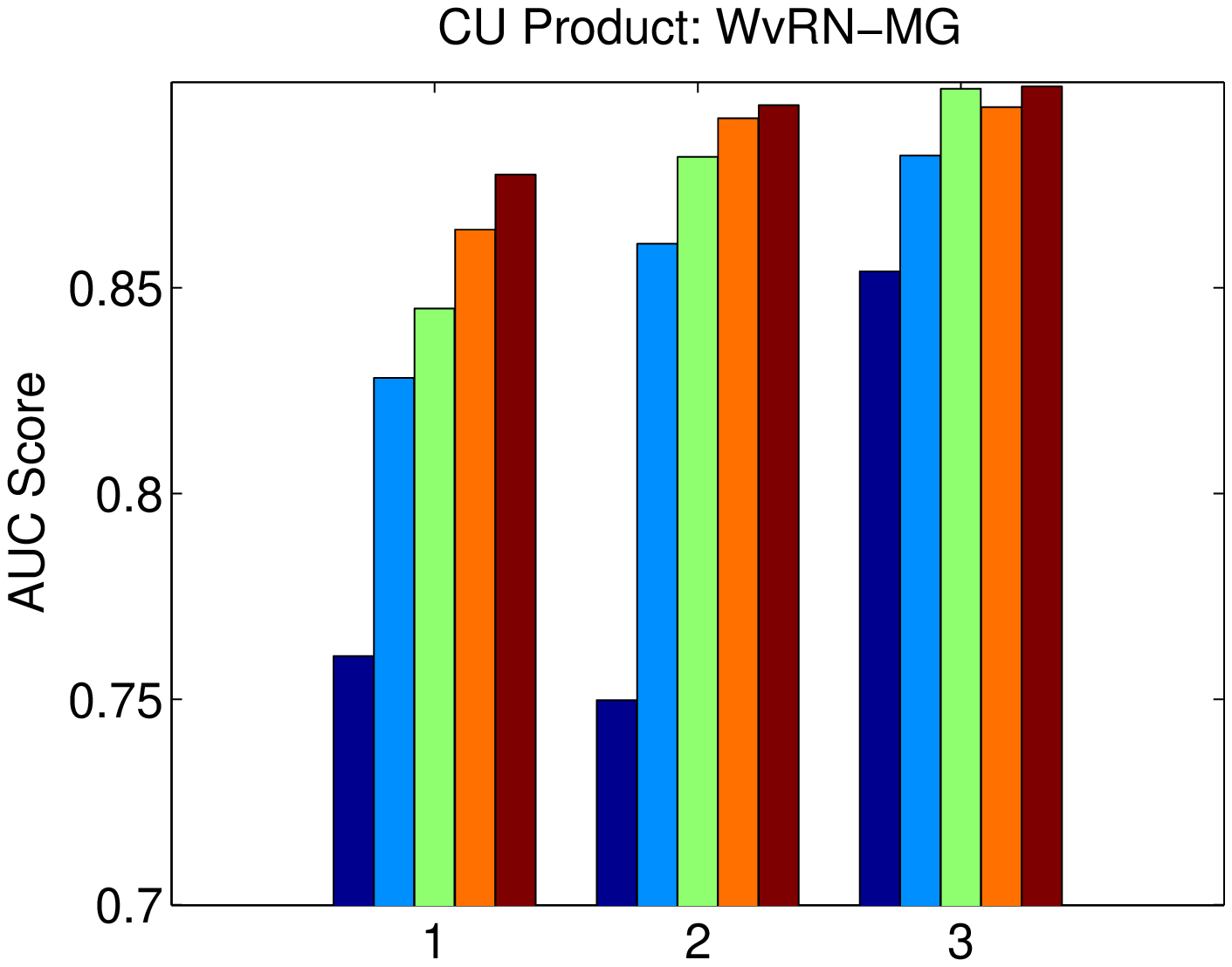} 
        \includegraphics[width=3.5cm,height=3.5cm]{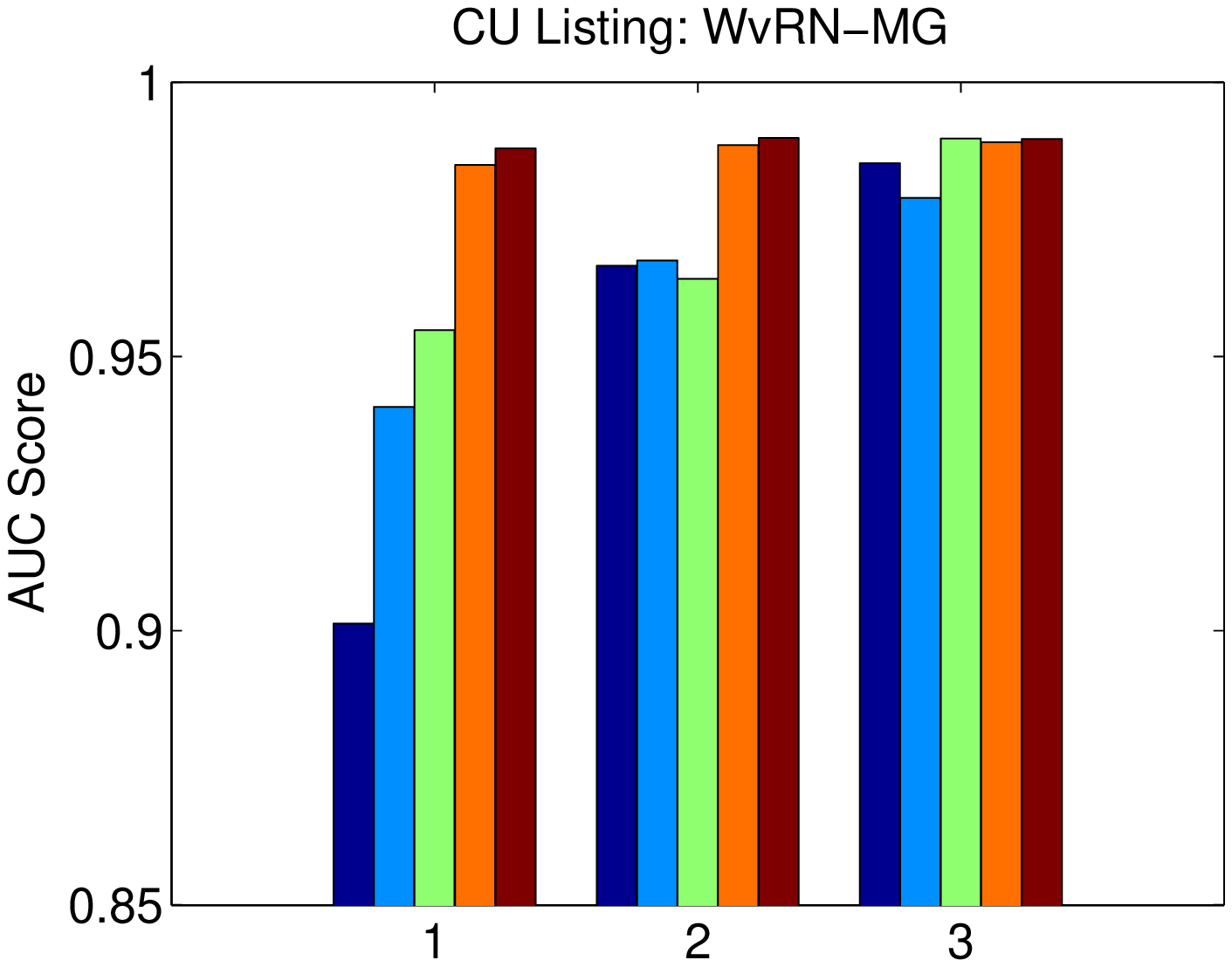} \\
        \includegraphics[width=3.5cm,height=3.5cm]{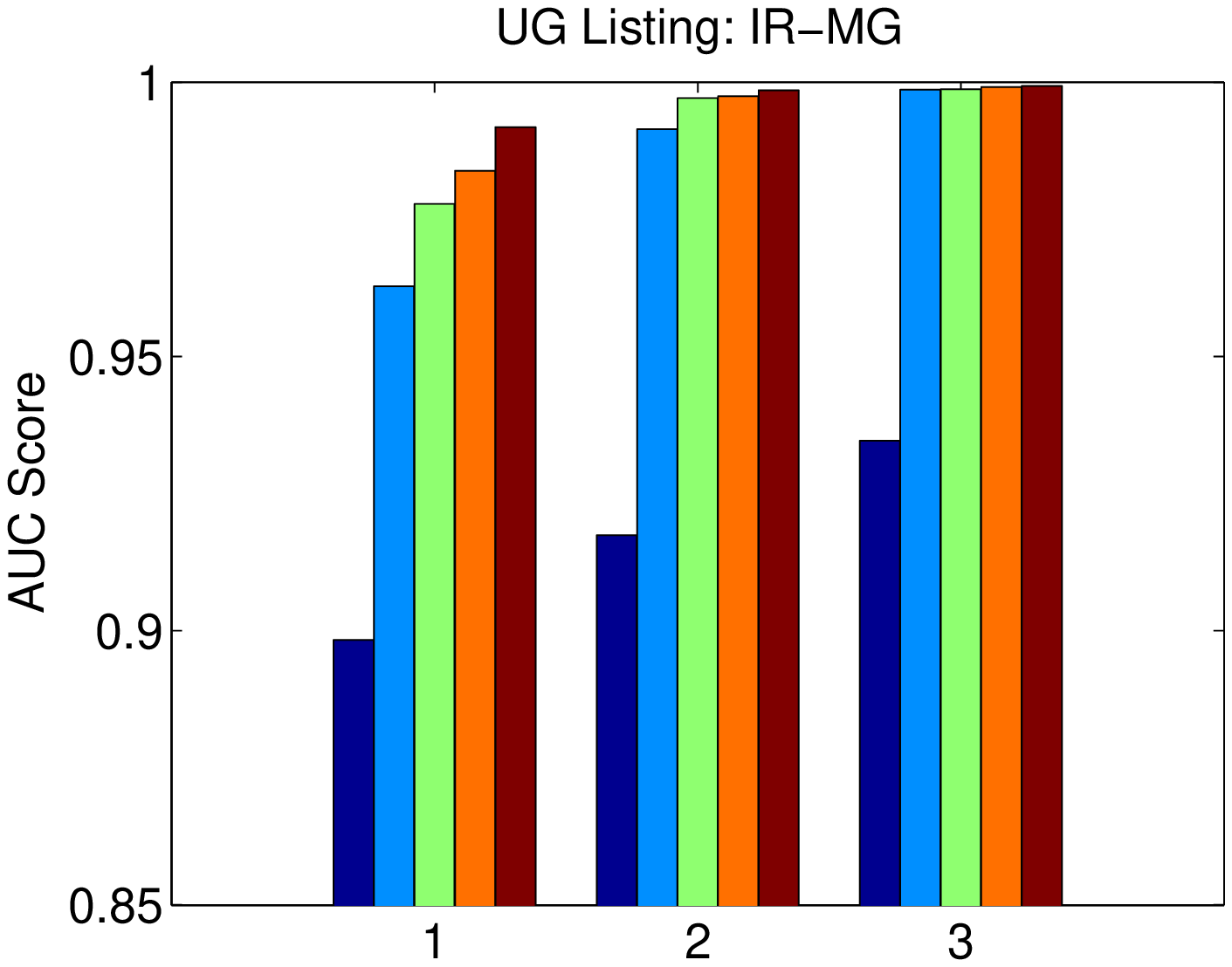} 
        \includegraphics[width=3.5cm,height=3.5cm]{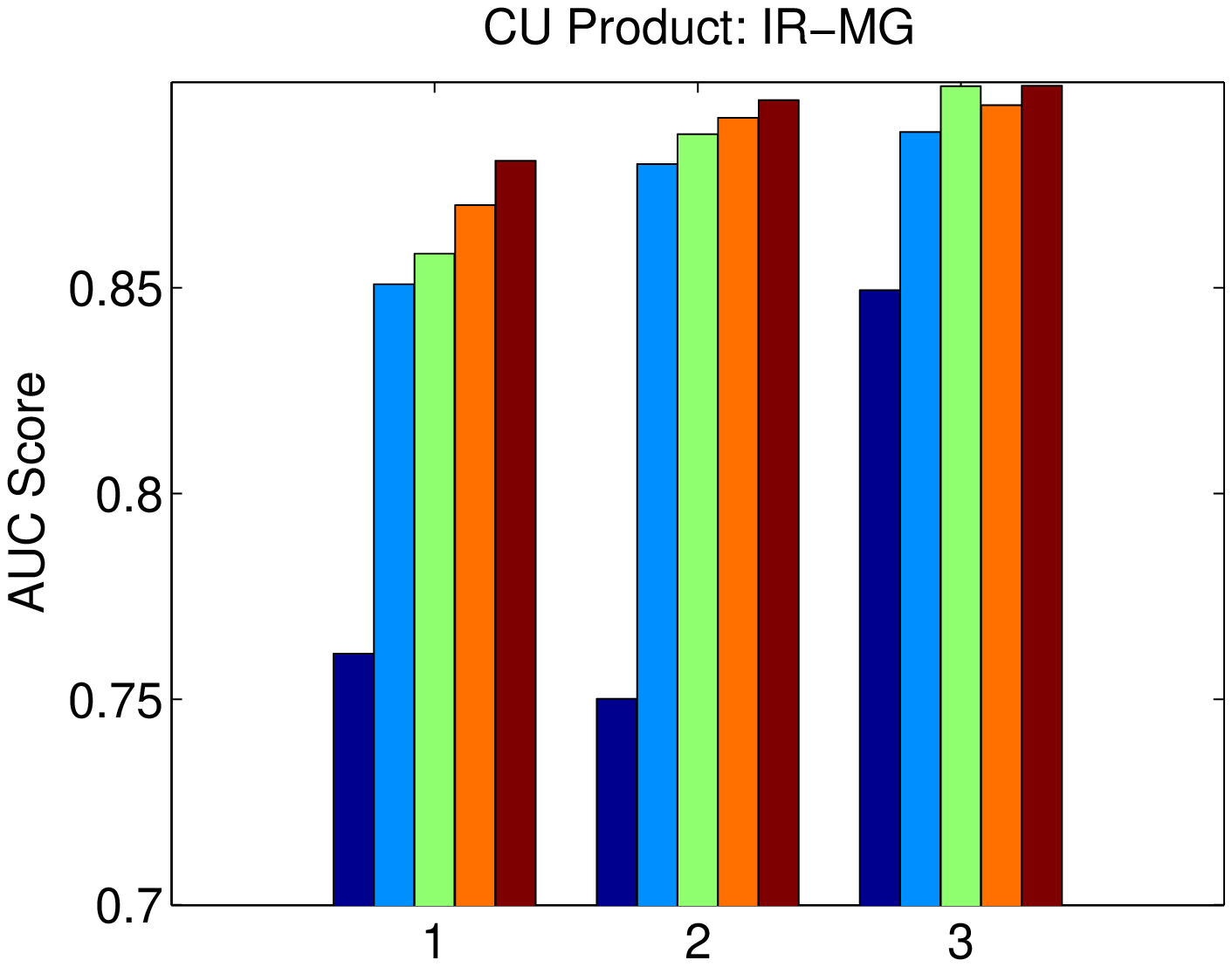} 
        \includegraphics[width=3.5cm,height=3.5cm]{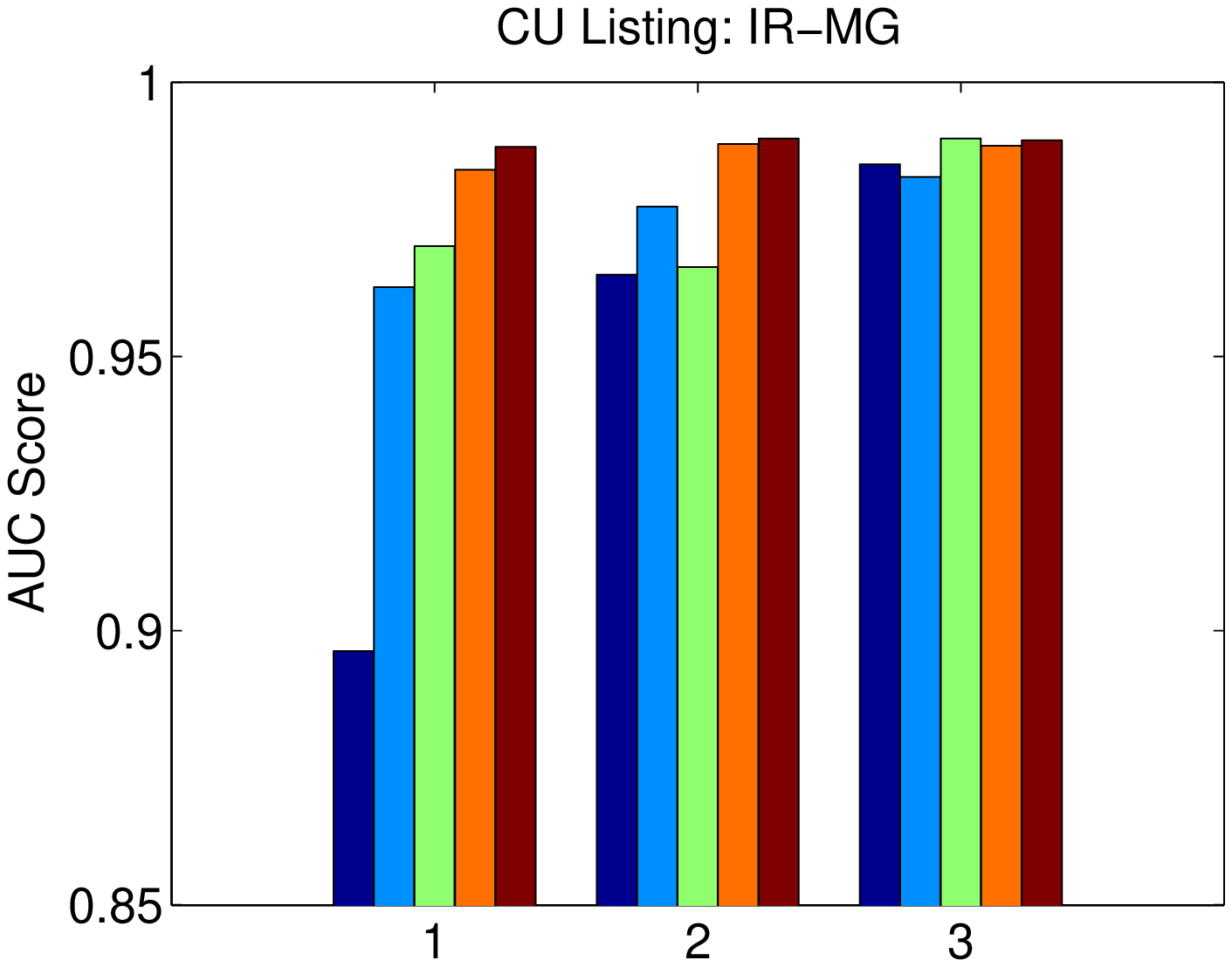} 
\caption{{AUC score performance of the IR-MG and WvRN-MG methods on two shopping domain datasets under three different label sizes (40, 80 and 160 - indicated as 1, 2 and 3) for dissimilar (dark blue), similar (blue) and mixed graph (3 cases - with NAC (green), CV (orange) and the best (maroon) $\gamma$ values) (in that order). 
}} \label{fig2}
\end{figure*}

\begin{table*}[t]
\begin{center}
\vspace{0.1in}
\caption{AUC Performance comparison of Goldberg et al., IR-MG and WvRN-MG methods on various datasets. The number of labeled examples (L) used in each dataset is indicated in parentheses. The number of realizations in each case was 25. The $\gamma$ values used in IR-MG and WvRN-MG are indicated in parentheses - here, NAC and CV indicate the techniques that were used to set $\gamma$.}
\label{tab:perfcomp2}
\vskip 0.05in
\begin{tabular}{|l|r|r|r|r|}\hline
Dataset & \multicolumn{1}{|c} {Method} & \multicolumn{1}{|c}{$P=5$} & \multicolumn{1}{|c}{$P=10$} & \multicolumn{1}{|c|}{$P=20$} \\ \hline
G50C (50)  &  WvRN-MG (NAC) & 0.9844 $\pm$ 0.0043  & 0.9886 $\pm$ 0.0040 & 0.9983 $\pm$ 0.0014 \\
  &  WvRN-MG (CV) & 0.9916 $\pm$ 0.0042  & 0.9930 $\pm$ 0.0061 & 0.9970 $\pm$ 0.0043 \\
  &  IR-MG (NAC) & 0.9851 $\pm$ 0.0042  & 0.9892 $\pm$ 0.0039 & 0.9986 $\pm$ 0.0012 \\
  &  IR-MG (CV) & 0.9914 $\pm$ 0.0055  & 0.9938 $\pm$ 0.0059 & 0.9967 $\pm$ 0.0048 \\
  &  Goldberg et al. & 0.9886 $\pm$ 0.0016  & 0.9946 $\pm$ 0.0011 & 0.9980 $\pm$ 0.0007 \\ \hline
WINDOWSMAC (100)  &   WvRN-MG (NAC) & 0.9632 $\pm$ 0.0056  &   0.9714 $\pm$ 0.0050   &   0.9927 $\pm$ 0.0026 \\
  &  WvRN-MG (CV) & 0.9811 $\pm$ 0.0091  & 0.9887 $\pm$ 0.0084 & 0.9938 $\pm$ 0.0061 \\
  &  IR-MG (NAC) & 0.9639 $\pm$ 0.0056  & 0.9722 $\pm$ 0.0050 & 0.9933 $\pm$ 0.0024 \\
  &  IR-MG (CV) & 0.9815 $\pm$ 0.0090  & 0.9883 $\pm$ 0.0082 & 0.9940 $\pm$ 0.0015 \\
  &  Goldberg et al. & 0.9714 $\pm$ 0.0029  & 0.9863 $\pm$ 0.0012 & 0.9950 $\pm$ 0.0003 \\ \hline
WebKB-LINK (40)  &   WvRN-MG (NAC) & 0.9465 $\pm$ 0.0120  &   0.9524 $\pm$ 0.0120  &   0.9696 $\pm$ 0.0074 \\
 &   WvRN-MG (CV) & 0.9626 $\pm$ 0.0073  &   0.9723 $\pm$ 0.0059  &   0.9800 $\pm$ 0.0041 \\
 &   IR-MG (NAC) & 0.9432 $\pm$ 0.0113  &   0.9499 $\pm$ 0.0118  &   0.9693 $\pm$ 0.0074 \\
 &   IR-MG (CV) & 0.9614 $\pm$ 0.0077  &   0.9718 $\pm$ 0.0062  &   0.9801 $\pm$ 0.0042 \\
 &   Goldberg et al. & 0.9451 $\pm$ 0.0260  &   0.9545 $\pm$ 0.0230  &   0.9607 $\pm$ 0.0201 \\ \hline
\end{tabular}
\end{center}
\end{table*}
\subsection{Evaluation on natural graphs from shopping sites}
We also evaluated the proposed methods on natural graphs constructed using structural signature (shingle) of web pages from shopping sites \url {http://www.uncommongoods.com} (referred as UG) and \url {http://www.compusa.com} (referred as CU). The similar and dissimilar graphs were constructed as follows. A similar edge between two pages was formed when their structural signatures had a match score of at least 6 (the values are in the range [0,8]) and, a dissimilar edge was put when the match score was 0\footnote{We used binary representation (i.e., edge with unit weight or no edge) for the graphs since the signatures are not accurate.}. In practice both the dissimilar and similar graphs have noise since the signatures are not accurate. We considered two binary classification problems. In the first problem, the goal was to differentiate {\it product detail} pages from the rest. In the second problem, the intent was to distinguish {\it product listing} pages from others. The properties of the datasets are given in Table \ref{tab:properties}.

Since the similar and dissimilar graphs are fixed we varied only the number of labeled nodes ($L = 40, 80, 160$). We evaluated the AUC performance of the IR-MG and WvRN-MG methods on the similar graph ($\gamma=1$) and dissimilar graph ($\gamma=0$) separately. Further, we evaluated the performance on the mixed graph for the values of $\gamma$ set by the NAC and CV based estimation techniques. To study the quality of these two estimation techniques, we also found the best AUC score given by the optimal $\gamma$ (searched over a grid of $\gamma$ values in the interval $[0,1]$ used in the cross-validation). The average performance over 25 partitions for each of these settings is presented in figure 2. It is clearly seen that the performance with the dissimilar graph is inferior compared to the performance with the similar graph, particularly when $L$ is small. This correlates well with the NAC values given in Table \ref{tab:properties}. Although the dissimilar graph is quite impure, it is still useful. This is clearly seen in figure 2 where the performance with the mixed graph is better than the performance with similar and dissimilar graphs used alone; see for instance the results for CU-Listing, WvRN, L=40. This improvement is quite significant when $L$ is small. Further, the performance with the cross-validation choice of $\gamma$ is very close to the best performance and is only slightly inferior when $L=40$. The NAC based estimate of $\gamma$ becomes useful for sufficiently large values of $L$. The performance difference between the IR-MG and WvRN-MG methods was statistically significant at the level of 0.05 only on the CU-Product and CU-Listing datasets when $L=40$. We have not reported the results for the UG-Product dataset since the AUC scores were almost same (around 0.99) for all the graphs and methods.

\subsection{Comparison with Goldberg et al.'s method}
Since Goldberg et al.'s method \cite{gold} depends on content features we restrict our comparison to the four datasets, G50C, WINDOWSMAC, WebKB-PAGELINK and WebKB-LINK. Goldberg et al. give two methods: one is based on regularized least squares (Lap-RLSC) and the other is based on SVMs (Lap-SVM) \cite{sindhwani}. Both methods perform similarly. We use Lap-RLSC for comparing against IR-MG and WvRN-MG. For IR-MG and WvRN-MG we tuned $\gamma$ using both cross validation (CV) and NAC values; CV tuning is obviously better and it is the one that should be used. The results for the methods are given in Table \ref{tab:perfcomp2} for various values of $P$. Clearly all three methods give competitive performance. The results are statistically significant for lower values of $P$. As in~\cite{gold}, for Goldberg et al.'s method we did not tune the hyperparameters for each choice of $P$. In the next section we show how tuning can be done and demonstrate its usefulness. In terms of computational speed Goldberg et al.'s method is comparable with IR-MG; WvRN-MG has an advantage over the other two methods because it is much faster ($>$ 10 times) and also provides decent competitive performance. 
\subsection{Setting up $\gamma$ parameter in Goldberg et al.'s method}
The above experiments clearly indicate the importance of $\gamma$ in the mixed graph to get improved performance. It would be useful to introduce such a parameter in Goldberg et al.'s method \cite{gold} also. One way of doing this is as follows. In their method there is a graph regularization term ${\bf f}^T{\bf M}{\bf f}$ which smoothens the decision function. Here, ${\bf f}$ corresponds to a vector of function values at the nodes of the graph and the matrix ${\bf M}$ is a mixed graph analog of the graph Laplacian ${\bf L}$. The combinatorial graph Laplacian matrix ${\bf L}$ is defined as ${\bf L}={\bf D}-{\bf W}$ where ${\bf D}$ is the diagonal degree matrix with $D_{ii}=\sum_{j=1}^n w_{ij}$ and its normalized version is given as: ${\bf L}_N={\bf I}-{\bf D}^{-{1\over 2}}{\bf W}{\bf D}^{-{1 \over 2}}$. ${\bf M}$ is defined as: ${\bf M}={\bf L}+({\bf 1}-{\bf J})\bullet{\bf W}$ where ${\bf 1}$ is a matrix of all ones and $\bullet$ is the Hadamard (elementwise) product. ${\bf J}$ is an edge type matrix with (i,j)~th element $J_{ij}=1$ if there is a similarity edge between $i,j$; $J_{ij}=-1$ if there is a dissimilarity edge. To introduce $\gamma$ we can modify ${\bf M}$ to be a convex combination of matrices ${\bf M}_S$ and ${\bf M}_{\bar S}$ corresponding to the similar and dissimilar graphs; that is, we set ${\bf M}=\gamma{\bf M}_S+ (1-\gamma){\bf M}_{\bar S}$. Using convex combination of Laplacian has been studied~\cite{sindhwani2} in the context of multiview learning. Here, ${\bf M}_S$ is nothing but the graph Laplacian ${\bf L}_{S}$ obtained using ${\bf W}_{S}$ and ${\bf M}_{\bar S}={\bf L}_{\bar S}+2{\bf W}_{\bar S}$. To verify the usefulness of this we conducted a simple experiment on the LINK dataset by setting $\gamma$=0.7, $P$=1.0 and $L$=20.  While the original method gave an average AUC score of 0.93, the modified method gave a value of 0.96. Like earlier, $\gamma$ can be tuned using cross-validation along with the other hyperparameters.
\section{Conclusion}
In this paper we provided a principled approach to extend probabilistic scores based transductive classification methods for mixed graphs. The proposed methods are simple and efficient. We highlighted the importance of hyperparameter optimization and showed how this parameter can be optimized particularly when the number of labeled nodes is not too small. Experiments on several benchmark and real world datasets show the usefulness of the proposed methods.  

\section{Acknowledgments}

The authors are thankful to the anonymous reviewers for their helpful comments. 

\bibliographystyle{abbrv}
\bibliography{DisGraph}

\begin{thebibliography}{10}

\bibitem{cord}
A.~Corduneanu and T.~Jaakkola.
\newblock Distributed information regularization on graphs.
\newblock In {\em NIPS}, pages 297--304, 2005.

\bibitem{cover}
T.~Cover and J.~Thomas.
\newblock {\em Elements of Information Theory}.
\newblock Wiley, 1991.

\bibitem{gold}
A.~B. Goldberg, X.~Zhu, and S.~Wright.
\newblock Dissimilarity in graph-based semi-supervised classification.
\newblock In {\em AISTATS}, 2007.

\bibitem{sofusaaai}
S.~A. Macskassy.
\newblock Improving learning in networked data by combining explicit and mined
  links.
\newblock In {\em AAAI}, 2007.

\bibitem{sofusjmlr}
S.~A. Macskassy and F.~Provost.
\newblock Classification in networked data: A toolkit and a univariate case
  study.
\newblock {\em JMLR}, 8:935--983, 2007.

\bibitem{newman}
M.~E.~J. Newman.
\newblock Mixing patterns in networks.
\newblock In {\em Physical Review E}, 2003.

\bibitem{sindhwani}
V.~Sindhwani, P.~Niyogi, and M.~Belkin.
\newblock Beyond the point cloud: from transductive to semi-supervised
  learning.
\newblock In {\em ICML}, 2005.

\bibitem{sindhwani2}
V.~Sindhwani, P.~Niyogi, and M.~Belkin.
\newblock A co-regularization approach to semi-supervised learning with
  multiple views.
\newblock In {\em ICML Workshop on learning with multiple views}, 2005.

\bibitem{subra}
A.~Subramanya and J.~Bilmes.
\newblock Entropic graph regularization in non-parametric semi-supervised
  classification.
\newblock In {\em NIPS}, 2009.

\bibitem{tong}
W.~Tong and R.~Jin.
\newblock Semi-supervised learning by mixed label propagation.
\newblock In {\em AAAI}, 2007.

\bibitem{zhou}
D.~Zhou, O.~Bousquet, T.~N. Lal, J.~Weston, and B.~Scholkopf.
\newblock Learning with local and global consistency.
\newblock In {\em NIPS}, pages 321--328, 2004.

\bibitem{zhu}
X.~Zhu, Z.~Ghahramani, and J.~Lafferty.
\newblock Semi-supervised learning using gaussian fields and harmonic function.
\newblock In {\em ICML}, pages 912--919, 2003.

\end{thebibliography}
\end{document}